# A Multi-Level Hierarchical Framework for the Classification of Weather Conditions and Hazard Prediction

Harish Neelam[1]

[1]Department of Epidemiology and Biostatistics, Michigan State University, USA
E-mail: neelamha@msu.edu

**Abstract**
*This paper presents a multi-level hierarchical framework for the classification of weather conditions and hazard prediction. In recent years, the importance of data has grown significantly, with various types like text, numbers, images, audio, and videos playing a key role. Among these, images make up a large portion of the data available. This application shows promise for various purposes, especially when combined with decision support systems for traffic management, afforestation, and weather forecasting. It's particularly useful in situations where traditional weather predictions are not very accurate, such as ensuring the safe operation of self-driving cars in dangerous weather. While previous studies have looked at this topic with fewer categories, this paper focuses on eleven specific types of weather images. The goal is to create a model that can accurately predict weather conditions after being trained on a large dataset of images. Accuracy is crucial in real-life situations to prevent accidents, making it the top priority for this paper. This work lays the groundwork for future applications in weather prediction, especially in situations where human expertise is not available or may be biased. The framework, capable of classifying images into eleven weather categories—dew, frost, glaze, rime, snow, hail, rain, lightning, rainbow, and sandstorm—provides real-time weather information with an accuracy of 93.29%. The proposed framework addresses the growing need for accurate weather classification and hazard prediction, offering a robust solution for various applications in the field.*

*Keywords:*
*Weather, classification, hierarchical model, multi-level classification, hazard prediction, VGG16.*

## 1. INTRODUCTION

In recent years, the increasing importance of data, particularly images, has been notable. Integrating weather image classification into decision support systems for tasks like traffic management, afforestation, and weather forecasting shows great potential, especially in scenarios where traditional forecasts are insufficient. While previous research has touched on these areas [1]-[3], this project uniquely concentrates on eleven distinct classes of weather images. Its principal aim is to construct a highly accurate model, trained on a sizable dataset, for predicting weather conditions, with a focus on precision to mitigate potential risks.

The main concerns in the domain of computer vision include object recognition, segmentation, and classification of images. An ML model must be capable of classifying the image among the other types before it can predict the image. Advanced techniques like machine learning and deep learning should be implemented to efficiently analyze the images that are obtained by sensors and cameras. Machine learning techniques like deep learning can be used to find features in imagery [4]. It makes use of a neural network, a multi-layered computer architecture created to resemble the functioning of the human brain. Each layer can extract one or more unique characteristics from an image.

This study involves more than 6000 images with 11 different types, it is important to capture every event in the image to accurately classify them. Hence, Deep learning techniques were very useful in this case to extract the features of an image [5]. Figure 1 describes the deep learning architecture with Input, few hidden layers and output of 11 classes.

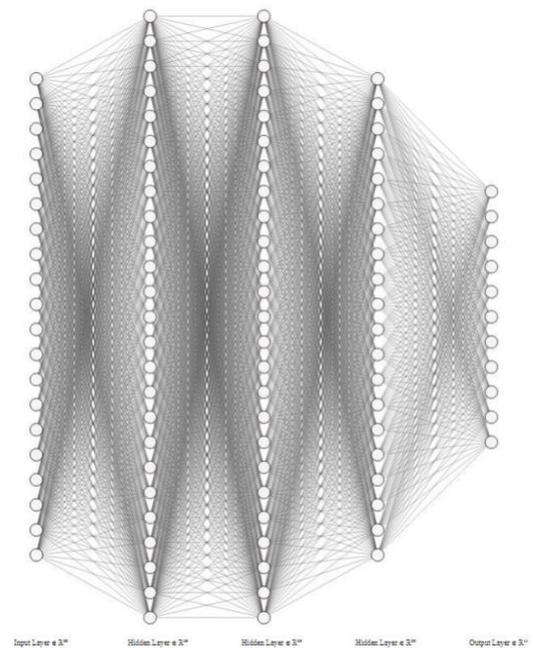

Fig.1. Deep learning Architecture

The project workflow includes several important stages: acquiring, analyzing, and preprocessing the dataset; developing and training machine learning models; and finally, selecting the most effective model.

In summary, this project aims to use machine learning, particularly deep learning, for accurate weather classification and hazard prediction. By using a multi-level hierarchical framework, it addresses the complexities of weather classification and sets the stage for better applications in weather forecasting and hazard mitigation.

## 2. DATA ANALYSIS AND PREPROCESSING

### 2.1 DATA SET

The data obtained has 6862(~7K) images of eleven classes (dew, fog/smog, frost, glaze, hail, lighting, rainbow, rain, rime, sandstorm, snow) published by Harvard Dataverse in 2021 [6]. The images of these classes are shown in Figure 2.

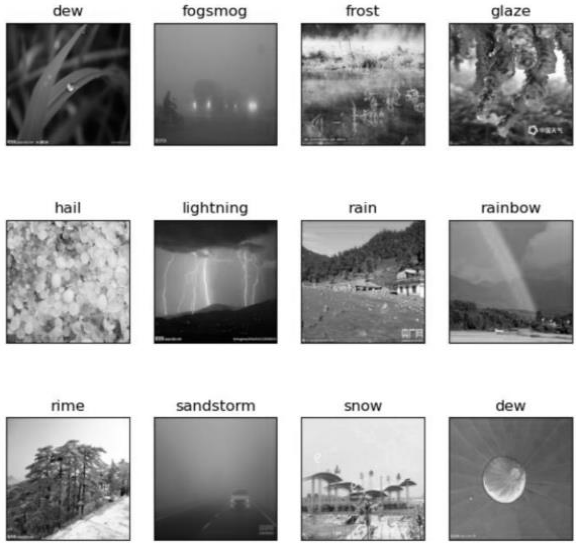

Fig 2. Sample Images in the Data set

### 2.2 EXPLORATORY DATA ANALYSIS

The Dataset was explored using Matplotlib and Seaborn to identify any patterns or trends in the data [7]. Figures 3 illustrates the distribution of target variables in the data.

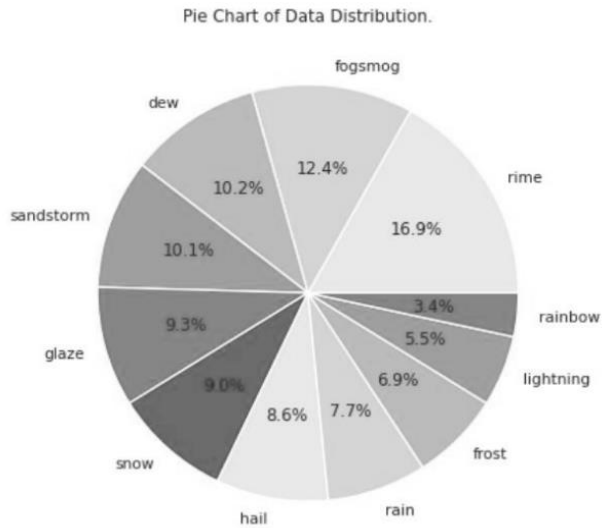

Fig 3. Class Distribution in the Dataset

The Data was then split into Training and Testing sets(70% Training data and 30% Testing data). The Training data was further split into training and validation data with a ratio of 80:20. The validation set is used to check the performance of the model while training with the training data to improve its performance. Figure 4 shows the distribution of data in training and testing data respectively to see the class distribution.

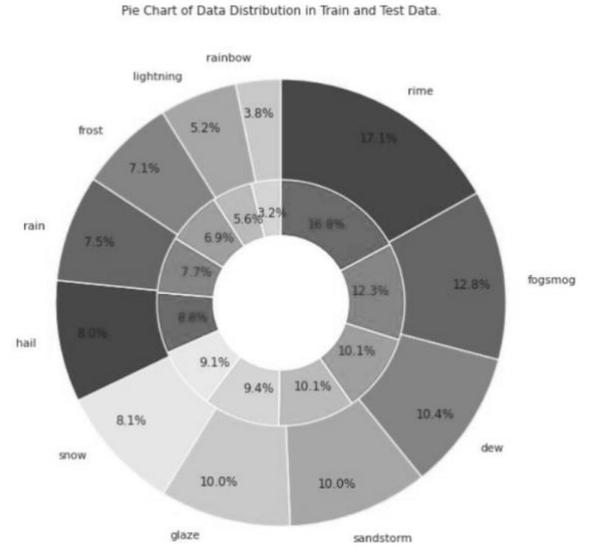

Fig 4. Nested Pie Chart of Class distribution in training and testing data

### 2.3 IMAGE PREPROCESSING

**Resizing:** Resizing refers to the process of changing the size of an image. It is crucial to choose an appropriate interpolation method while resizing an image to preserve its quality as much as possible. In this case, the images were resized to 100x100x3.

**Interpolation:** In Image processing, interpolation is used to create new pixels when resizing an image to a different size. Lanczos interpolation is used to resize the images to 100x100x3, because it is considered a good method for image resizing as it helps to preserve image quality and sharpness. Unlike other interpolation methods, Lanczos interpolation uses a more complex formula to estimate the values of new pixels based on the surrounding pixels [8].

**Color Space Conversion:** Different color spaces represent colors in different ways, and some color spaces are better suited for specific applications than others. For example, the RGB color space is commonly used for displaying images on screens. There is evidence that RGB color spaced images train well, so the images were converted to the RGB color space [9]. Figure 5 show the images before and after Image Preprocessing.

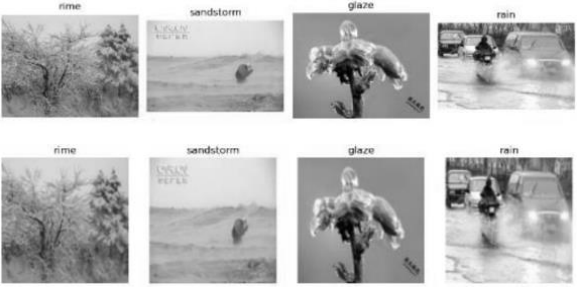

Fig 5. Images before and after preprocessing

## 2.4 NORMALIZATION

Each image is normalized before training the model. Normalization can help to improve the performance and accuracy of the machine learning model by ensuring that the input data is on a consistent scale and distribution [9], [10]. It also helps to prevent the influence of any outliers or extreme values in the data. The images were converted to lists, and the mean and standard deviation of the training set were calculated as shown in the Eq. (1). Each image array was then subtracted by the mean and divided by the standard deviation of the train data. These mean and standard deviation values were stored to be used during prediction.

$$X = \frac{X - X\_train\_mean}{X\_train\_std} \quad (1)$$

## 2.5 ONE-HOT ENCODING

Categorical data can be transformed into numerical values using the label encoding technique. In machine learning, algorithms typically require numeric input data, so label encoding is often used to preprocess the data before it can be used to train a model. In a categorical variable, label encoding gives each category a distinct integer value [11]. It is used to encode the Target variable. Binary vectors can be used to represent categorical variables using this technique, where each category is represented as a vector of all zeros except for one at the index corresponding to the category as shown in the following Eq. (2).

$$Class\ 0 = [1,0,0,0,0,0,0,0,0,0,0]$$
$$Class\ 3 = [0,0,0,1,0,0,0,0,0,0,0] \quad (2)$$

## 4. METHODS

## 4.1 SCIKIT-LEARN ESTIMATORS

The following are some of the Scikit-Learn estimators used for this data as baseline models. The initial data obtained has images with different sizes and labels. The images were resized to 32x32x3 and flattened to 3,072 pixels. The data now has 3,072-pixel features and label column as target variable. This data is used to perform the following estimators.

### 4.1.1 Random Forest

A set of decision trees that grow in a selection of randomly chosen data subspaces are called random forests and are a method for creating classification ensembles. According to experimental findings, random forest classifiers are capable of accurately classifying data in high-dimensional domains with a wide range of classes [12]. Recently, image classification and bioinformatics have shown increasing interest in random forests. The popularity and accuracy of this estimator made it the first option for this study to classify images. Unfortunately, this model did not perform well on this data. This was able to provide accuracy measure of 33.14% on average when ran multiple times. This may be because there were many features( 3,072pixels) and 11 classes. The accuracy was 50.60% when the target variable is grouped to 3 classes.

### 4.1.2 Support Vector Machine Classifier

Support vector machines are a category of supervised learning models for outlier identification, regression modeling, and data classification [13]. In high dimensional spaces, it performs very well. This was able to offer an accuracy of just 38.12% and took a huge amount of time to fit the model. Perhaps this is because there are many features and less samples, which suggests the necessity of deep learning models [14].

Deep learning models were proven to be the best in such cases where Scikit-Learn estimators doesn't perform well [5]. There are many pre-trained models and convolutional neural networks with various numbers of hidden layers. Some of them that are used are as follows:

## 4.1 BASIC CNN

A Basic CNN model is designed, and this model expects input images (100, 100, 3) size. The model architecture comprises of multiple hidden layers of convolutional, batch normalization, average pooling, and dropout layers, which are all used to extract important features from the input images and reduce the chances of overfitting [15]. The final layer is a dense layer with 11 output nodes, which corresponds to the number of classes that the model can predict. The activation function used for all the convolutional and dense layers, except the last one, is ReLU [16]. Using a SoftMax activation function, the final dense layer generates a probability distribution across the classes.

## 4.2 RESNET 50

The ResNet50 model is loaded with pre-trained weights on the ImageNet dataset and its last layer, which performs classification into the 1,000 ImageNet classes, is removed. The remaining convolutional layers in the ResNet50 model are set as non-trainable, so that only the dense layers added later will be trained [17]. Then, a Flatten layer is added to the output of the ResNet50 model to convert the 2D output tensor from the convolutional layers into a 1D vector. Two Dense layers are added on top of the flattened output. Finally, a model object is created, which takes the input of the ResNet50 model and outputs the probability distribution over the 11 classes generated by the two dense layers.

## 4.3 INCEPTIONV3

Google built a deep learning model called InceptionV3 for picture classification [18]. It has been demonstrated that InceptionV3 performs at the cutting edge on extensive picture classification datasets like ImageNet. InceptionV3 is also faster to train and more computationally efficient than some other deep learning models because it has fewer parameters overall. This makes it an excellent option for applications where speed and efficiency are important.

## 4.4 VGG16

The Visual Geometry Group (VGG) at the University of Oxford created the deep learning model VGG16 for classifying images. The accuracy of VGG16 in picture classification tasks is one of its key advantages. 13 convolutional layers and 3 fully linked layers make up the VGG16 architecture [15], [19] . To extract features from the input image, the convolutional layers utilize tiny 3x3 filters with a 1 stride and 1 padding. The fully connected layers output a probability distribution over the classes using a SoftMax activation function. Additionally, the design of VGG16 is quite uniform and straightforward to understand, making it simple to customize for particular applications. Due to its simplicity, it is also less likely to overfit smaller datasets. Since VGG16 has already been trained on large image classification datasets, it can be fine-tuned for specific applications using smaller datasets. This pre-training can enhance the accuracy and generalizability of the model. When fine-tuning VGG16 for a specific application, the existing pretrained weights can be used as a starting point [20]. By only training the final layers of the model on a smaller dataset, the model can be fine-tuned to identify specific patterns that are important for the target task. This is often more effective than training a new model from scratch because the pre-trained model already has a good understanding of basic image features.

## 4.5 HIERARCHICAL CLASSIFICATION MODEL

The images were divided into three groups based on the weather: Rainy, Dusty, and Cold. The images labeled as rain, hail, lightning, and rainbow were mapped to Rainy weather, while the images labeled as sandstorm and fog/smog were mapped to Dusty weather. All the other images were mapped to Cold weather. A model was built to first predict the main class of weather: **Rainy, Cold,** or **Dusty**. Based on this result, sub-models were trained using respective images to predict the type of weather and whether it is safe. This approach is called **Multi-level Hierarchical Classification Model** [21]. The first level of the model would classify the images into high-level groups, and the second level would classify sub-groups within each group.

## 5. RESULTS

The Basic CNN model gave an accuracy of 70.99%. The model was adjusted by increasing and decreasing the number of hidden layers, changing the number of filters, and adjusting the input shape. The best results were achieved with 26 hidden layers and an input shape of 100x100x3.

ResNet is a more complex architecture compared to basic CNN models and has demonstrated strong performance in many computer vision tasks. However, ResNet can sometimes underperform relative to a basic CNN model due to the risk of overfitting. The numerous parameters in ResNet can lead to overfitting if the model is not adequately regularized. The InceptionV3 architecture can be complex and difficult to interpret, which may make it harder to optimize and fine-tune for specific applications. On the other hand, basic CNN models have fewer parameters and are thus less susceptible to overfitting. This could explain why ResNet and InceptionV3 did not perform as well as the basic CNN model in this particular case [22].

VGG16 model performed very well on this data, achieving an accuracy of 80.38% compared with other models as shown in Table 1. This is notably higher than the performance of other models. Therefore, it was a good choice to proceed with this model for predicting the weather in the images, with some modifications.

Table 1. Testing accuracies of classification models

| Classification Model | Accuracy |
|---|---|
| **Random Forest** | 33.14% |
| **Support Vector Classifier** | 38.12% |
| **Basic CNN** | 70.99% |
| **ResNet50** | 55.95% |
| **InceptionV3** | 62.02% |
| **VGG16** | 80.38% |

## 5.1 HIERARCHICAL MODEL

**Multi-level Hierarchical Classification Model** was built to first predict the main class of weather: **Rainy, Cold,** or **Dusty**. Based on this result, sub-models were trained using respective images to predict the type of weather and whether it is safe. The Accuracy of the model that predicts Rainy, Dusty or Cold Weathers is **93.29%** and the confusion matrix is as shown in the Figure 6.

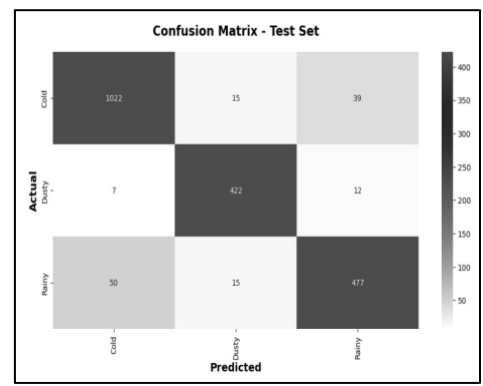

Fig 6. Confusion matrix of Primary model

It can be observed that when the model is trained with three classes, the accuracy is high [23]. The final model consists of a main primary model that classifies three labels (Rainy, Dusty, and Cold), and sub-models that classify Rainy Weather into four classes (Rain, Rainbow, Lightning, and Hail), Dusty Weather into two classes (Sandstorm and Fog/Smog), and Cold Weather in two ways. The first model classifies weather into five classes (Dew, Snow, Rime, Glaze, and Frost), while the other model classifies it into Safe or Potentially Hazardous weather. Therefore, the weather in the image is classified into eleven classes and into three classes (Safe, Potentially Hazardous, and Dangerous) [24]. The results of all the models are shown in Table 1.

The models that classify Rainy and Dusty weather produced accuracies of **95.17%** and **92.44%,** respectively. The model that classifies Cold weather into 5 classes gave an accuracy of about **79.85%.** The Confusion matrix of model classifying cold weather images is shown in the Figure 7.

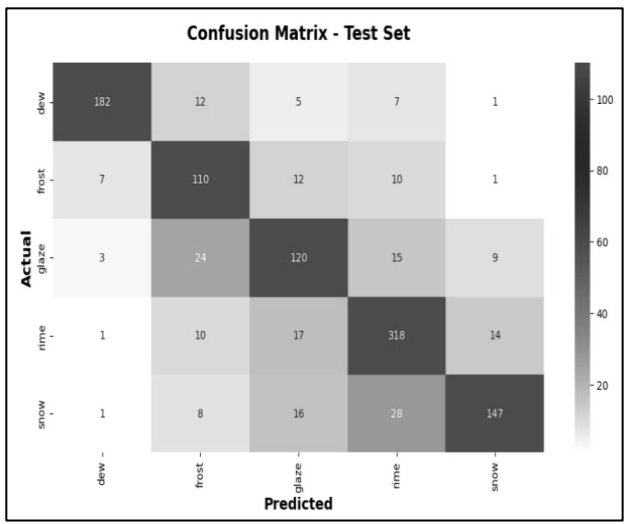

Fig 7. Confusion matrix of Secondary Cold weather model

From the results, it can be observed that glaze is often confused with frost, and snow is often confused with rime. Frost and glaze have very poor precision, possibly because the images of these two weathers resemble each other.

## 5.1 PREDICTION

The Best model, VGG16 [25], was chosen to perform the prediction on the image. The flow of the prediction is shown in Figure 8.

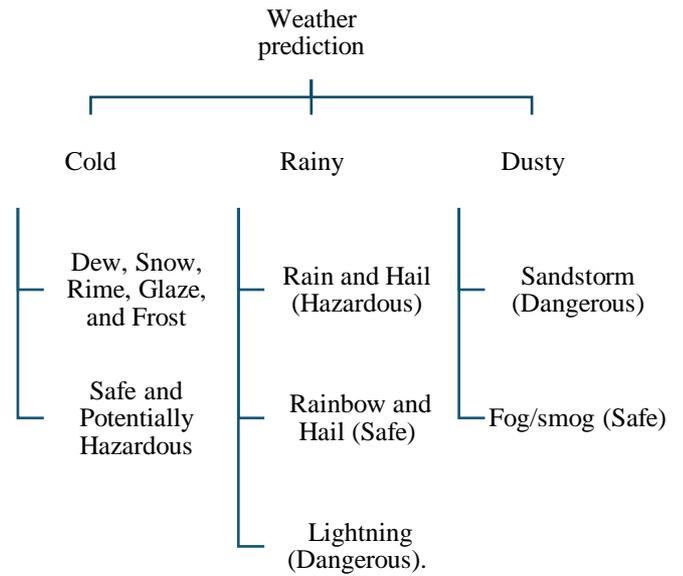

Fig 8. Flowchart of Hierarchical classification

## 6. CONCLUSION

This study aimed to develop a machine learning model capable of accurately classifying weather images into eleven distinct classes and assessing their safety levels as safe, potentially hazardous, or dangerous. Leveraging a dataset comprising thousands of images, the model's primary application is in weather forecasting and hazard prediction, especially in scenarios where human expertise is unavailable or potentially biased. The **VGG16** model emerged as the best performer, achieving notable accuracies, particularly for rainy and dusty weather conditions. The model's accuracy for cold weather classification was comparatively lower, primarily due to the visual similarities between certain weather types like frost and glaze.

By adopting a multi-level hierarchical classification framework, this paper effectively addressed the complexities involved in weather classification. The primary model classified images into three broad categories: Rainy, Dusty, and Cold. Subsequently, specialized sub-models further refined these categories into specific weather conditions and safety levels. This hierarchical approach demonstrated high accuracy, particularly for the primary weather categories and their associated sub-classes. The hierarchical model's ability to classify weather conditions accurately and its potential to enhance real-time weather forecasting and hazard prediction underscore its practical value. The application of this model could significantly mitigate risks associated with adverse weather conditions, providing a reliable tool for decision support in various domains such as traffic management and disaster preparedness.

In conclusion, this study emphasizes the efficiency of deep learning, particularly the VGG16 architecture, in complex weather classification tasks. The approach offers a robust foundation for future advancements in weather prediction applications, aiming to improve safety and preparedness in the face of varying weather conditions.

## ACKNOWLEDGMENTS

Dr. Michael Murillo, Professor of CMSE Department MSU.